\pdfoutput=1

\documentclass[11pt]{article}

\usepackage[]{acl}

\usepackage{times}
\usepackage{latexsym}

\usepackage[T1]{fontenc}

\usepackage[utf8]{inputenc}

\usepackage{microtype}
\usepackage{graphicx}
\usepackage{subfigure}
\usepackage{svg}
\usepackage{CJKutf8}
\usepackage{setspace}
\usepackage{multirow}
\usepackage{subfloat}
\usepackage{graphicx}
\usepackage{url}
\usepackage{tikz}
\usepackage{pgfplots}
\usepackage{color}
\usepackage{mathtools}
\usepackage{makecell}
\usepackage{colortbl,booktabs}
\usepackage{paralist}
\usepackage{amsmath,amssymb,bm}
\usepackage{threeparttable}

\title{SkillNet-NLU: A Sparsely Activated Model for General-Purpose Natural Language Understanding}
\author{Fan Zhang, Duyu Tang\thanks{\ \ \mbox{Correspondence to Duyu Tang (\mbox{duyutang@tencent.com}).}}, Yong Dai, Cong Zhou, Shuangzhi Wu \and Shuming Shi \\ \\  Tencent AI Lab}

\begin{document}
\maketitle
\begin{abstract}
Prevailing deep models are single-purpose and overspecialize at individual tasks. However, when being extended to new tasks, they typically forget previously learned skills and learn from scratch. We address this issue by introducing SkillNet-NLU, a general-purpose model that stitches together existing skills to learn new tasks more effectively. The key feature of our approach is that it is sparsely activated guided by predefined skills. Different from traditional dense models that always activate all the model parameters, SkillNet-NLU only activates parts of the model parameters whose skills are relevant to the target task. When learning for a new task, our approach precisely activates required skills and also provides an option to add new skills. We evaluate on natural language understandings tasks and have the following findings. First, with only one model checkpoint, SkillNet-NLU performs better than task-specific fine-tuning and two multi-task learning baselines (i.e., dense model and Mixture-of-Experts model) on six tasks. Second, sparsely activated pre-training further improves the overall performance. Third, SkillNet-NLU significantly outperforms baseline systems when being extended to new tasks.
\end{abstract}

\begin{CJK}{UTF8}{gbsn}

\section{Introduction}
Recent years have witnessed the success of homogeneous models based on  Transformer~\cite{vaswani2017attention} and pre-trained models~\cite{devlin2018bert} in artificial intelligence and natural language processing.
Many previous works use similar neural network models and repeat the same process: learning from scratch\footnote{In this work, the terminology ``from scratch'' refers to the unawareness of task knowledge, even if the model is initialized with pre-trained models like BERT~\cite{devlin2018bert}.} and fine-tuning all the model parameters for an isolated task. 
However, this differs from human learning in two aspects.
First, we human beings don't forget everything we have learned and start learning new skills from nothing. Instead, we combine existing skills to learn new skills faster.
Second, we have about 100 billion neurons in our brain and different parts are specialized for different skills. When we solve a problem, we don't activate all the neurons but only call on relevant parts.

\begin{figure*}[!t]
	\centering
	\subfigtopskip=2pt
	\subfigure[SkillNet-NLU for text classification. s1 and s7 activated.]{
	\includegraphics[width=0.47\linewidth]{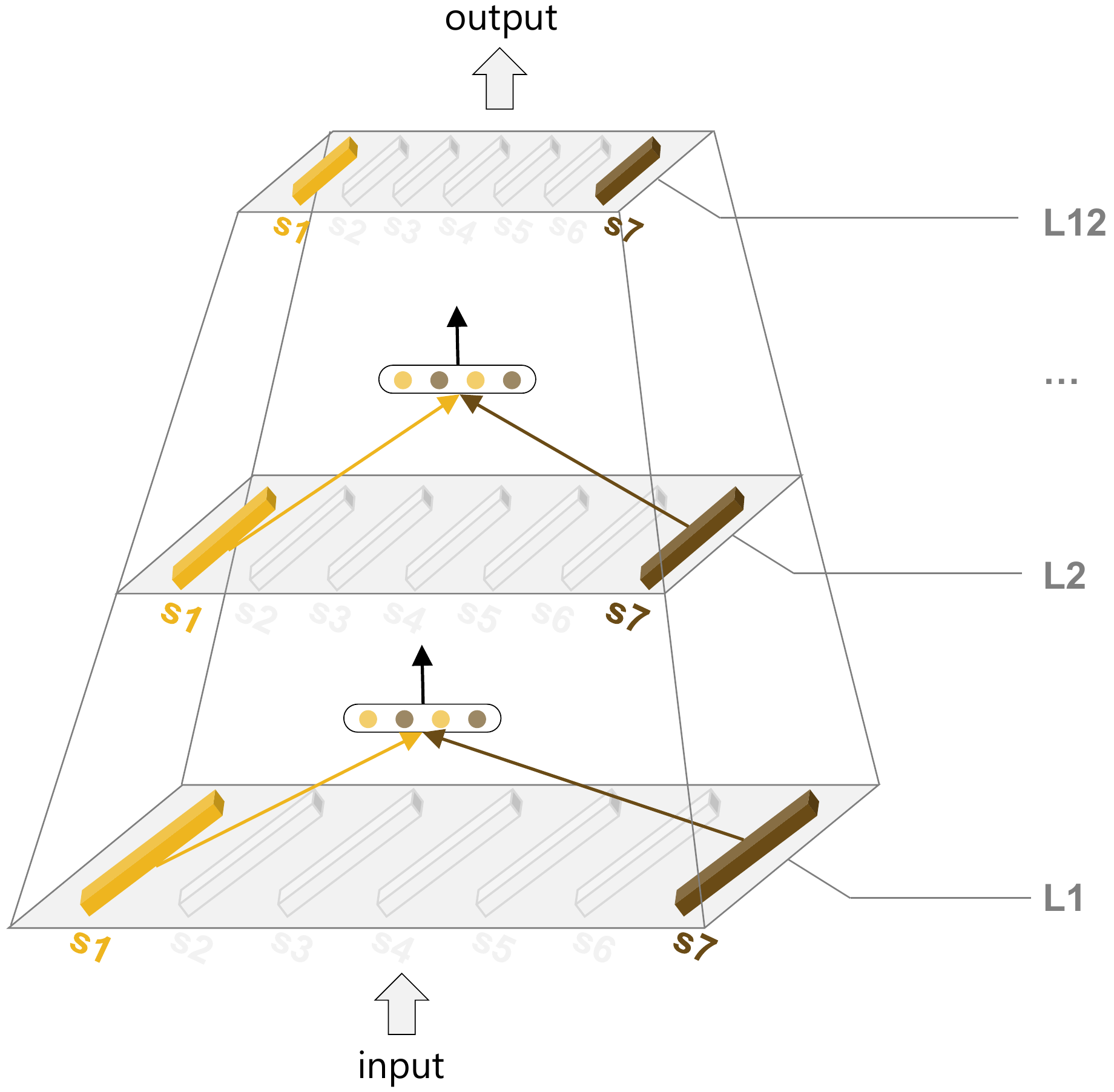}}\ \ \ \ \ \ \ \ 
	\subfigure[SkillNet-NLU for sentiment classification. s1, s4, s7 activated.]{
	\includegraphics[width=0.47\linewidth]{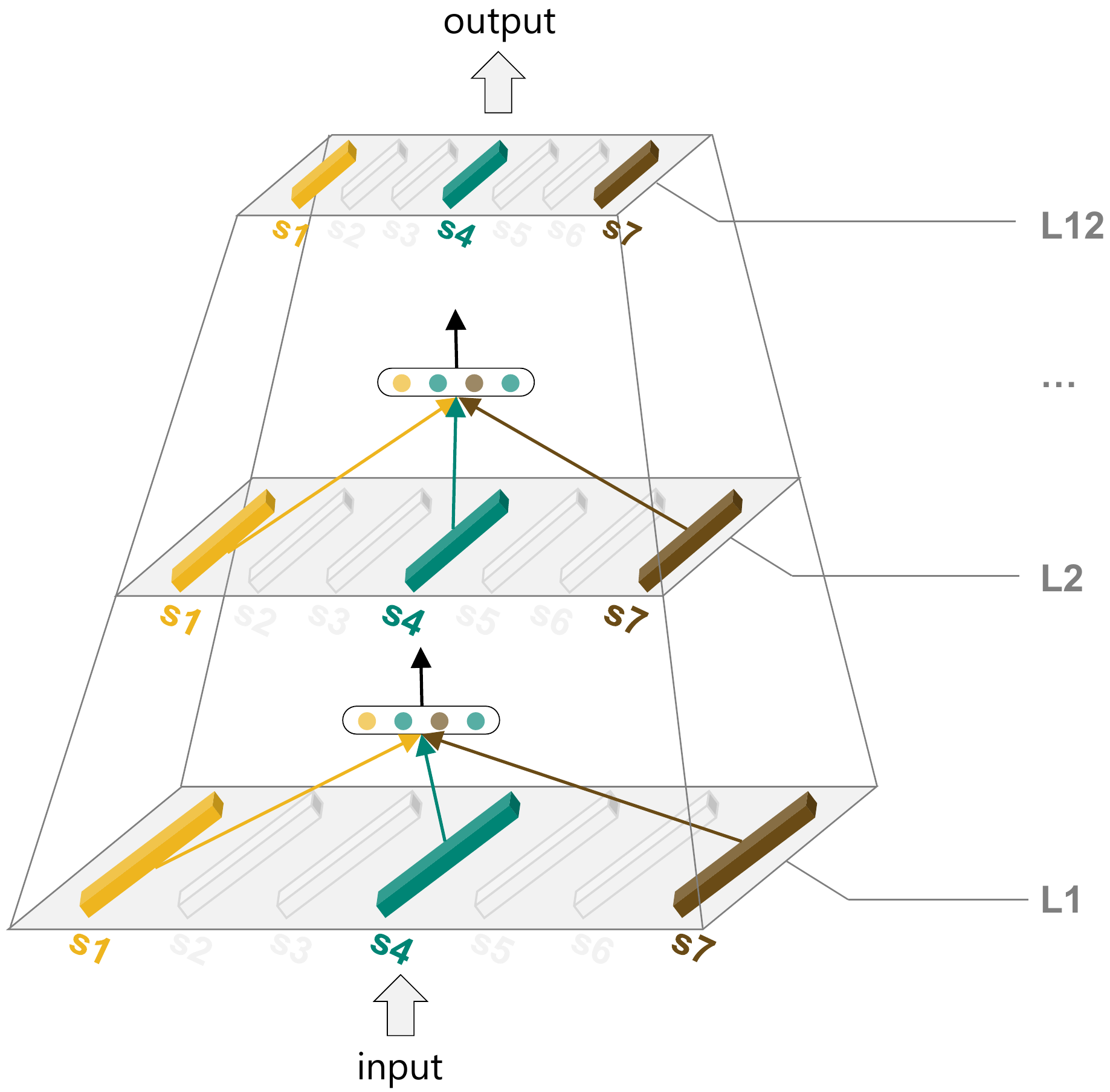}}
	
	\subfigure[SkillNet-NLU for machine reading comprehension. s2, s3, s5 and s7 activated.]{
	\includegraphics[width=0.47\linewidth]{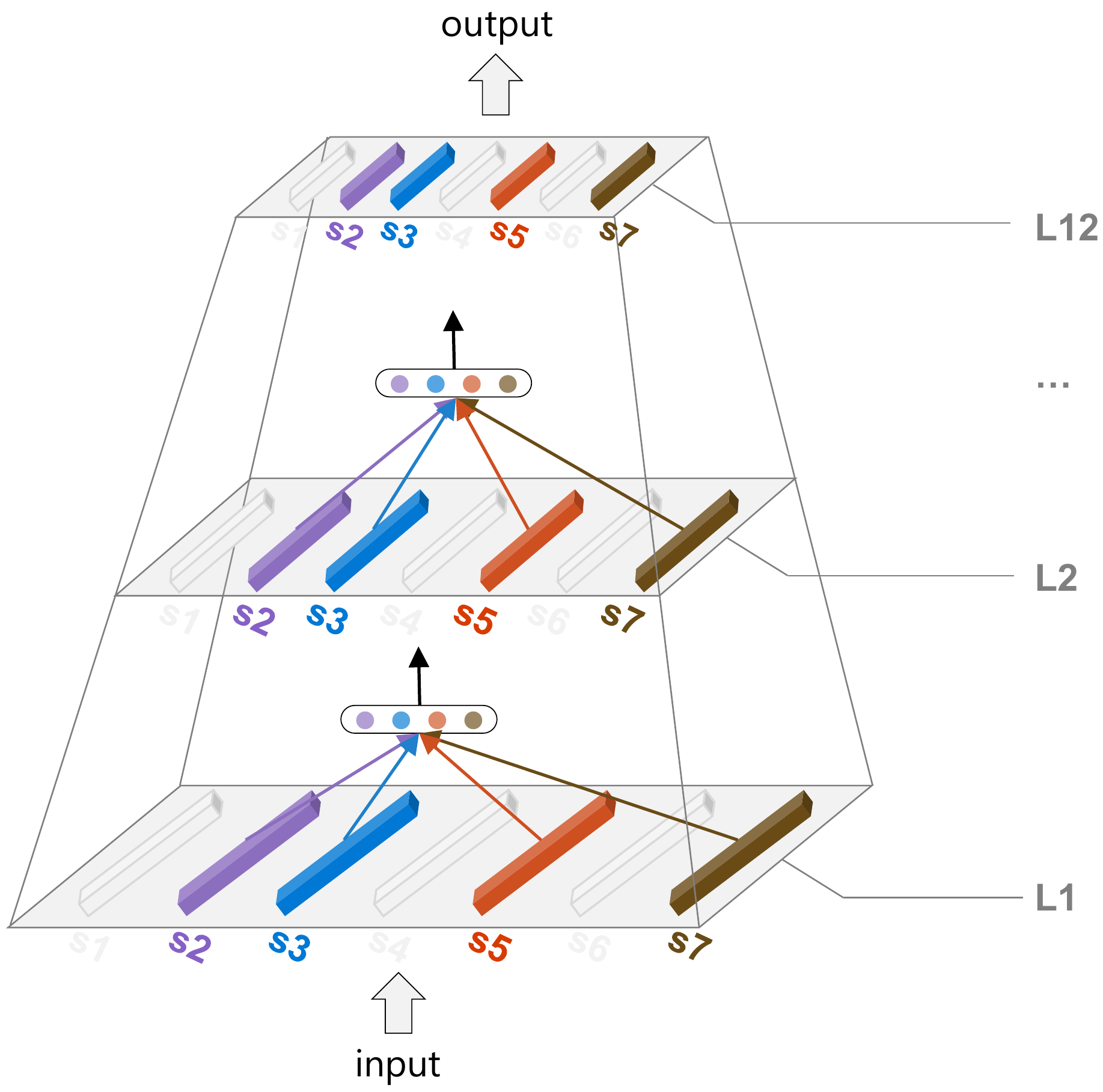}}\ \ \ \ \ \ \ \ 
	\subfigure[Mixture of experts.]{
	\includegraphics[width=0.47\linewidth]{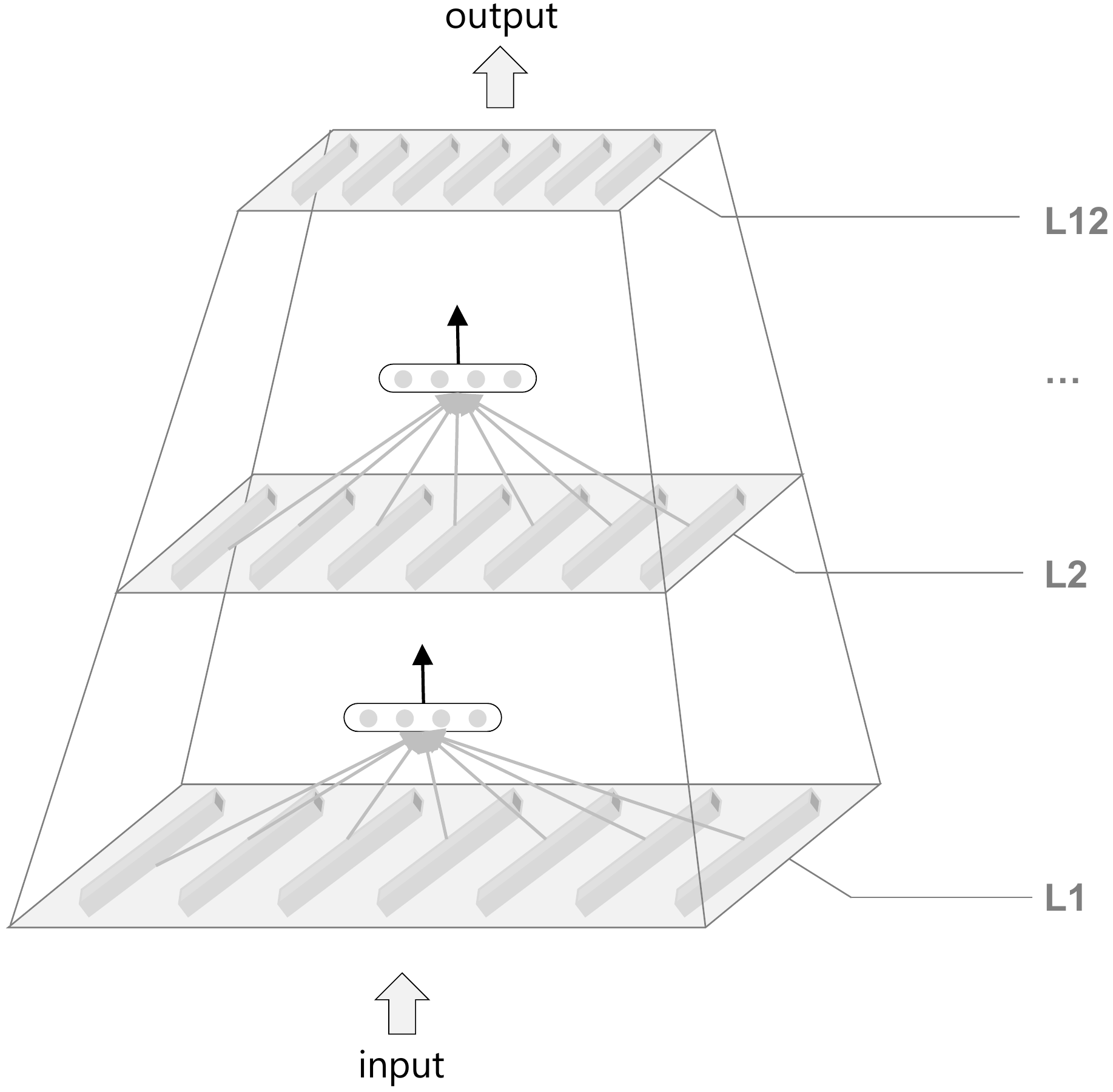}}
	\caption{Illustrative examples of our SkillNet-NLU for NLU tasks and the comparison to a fully activated MoE model. In SkillNet-NLU (a, b and c), each pillar is a skill module. Pillars filled in color (e.g., yellow, green, purple, blue, red and brown) are activated. Skills are defined in Table \ref{table:intro-skills}.}
	\label{fig:model-intro}
\end{figure*}

In this work, we present an approach to address the aforementioned issues.
Our goal is to advance from single-purpose models to general-purpose models and from dense models to sparse models.
Specifically, we take natural language understanding (NLU) as a case study and present a sparsely activated model that is capable of generalizing across many different NLU tasks.
The key feature of our approach is that it includes a set of reusable parameterized ``skill modules'', each of which corresponds to a skill such as \textit{the skill to understand the sentiment of texts}, \textit{the skill to understand natural language questions}, \textit{the skill to understand the meaning of texts in finance domain}, etc.
Different from traditional dense models that always activate all the model parameters, our approach sparsely activates parts of the model parameters, while deactivating the modules whose skills are irrelevant to the task.

Let's use three concrete examples to illustrate how our model is sparsely activated when it is adopted in downstream tasks. 
Let's suppose we have defined seven skills, whose definitions are given in Table \ref{table:intro-skills}. 
For the task of text classification, only the ability to get the semantic representation of a sequence (i.e., \texttt{s1}) is required. Therefore, only the parameters that relate to \texttt{s1} and \texttt{s7} are activated, as shown in Figure \ref{fig:model-intro} (a)\footnote{We define a generic skill \texttt{s7}, which is always activated as the default skill. This design aims to provide a backup for handling new tasks that require totally unseen skills.}.
Compared to text classification, sentiment classification requires an additional skill to understand the sentiment of texts. Therefore, \texttt{s1}, \texttt{s4} and \texttt{s7} are activated, as given in Figure \ref{fig:model-intro} (b).
For the task of machine reading comprehension, models need to understand the meaning of the question (\texttt{s5}), understand how question and passage interact (\texttt{s3}) and get the representation of each token (\texttt{s2}). Therefore, \texttt{s2}, \texttt{s3}, \texttt{s5} and \texttt{s7} are activated, as shown in Figure \ref{fig:model-intro} (c).

\begin{table}[t]
\centering
\begin{tabular}{ll}
\hline
\textbf{Skill} & \textbf{Description} \\
\hline
\verb|s1| & {get the semantic meaning of a sequence} \\
\verb|s2| & {get the semantic meaning of a token} \\
\verb|s3| & {understand how two text segments interact} \\
\verb|s4| & {understand the sentiment of texts} \\
\verb|s5| & {understand natural language questions} \\
\verb|s6| & {understand texts in finance domain} \\
\verb|s7| & {generic skill} \\
\hline
\end{tabular}
\caption{Examples of skills and descriptions.}
\label{table:intro-skills}
\end{table}
We briefly summarize how SkillNet-NLU differs from both multi-task learning methods and Mixture-of-Experts (MoE) methods as follows.

\begin{enumerate}
\item Multi-task learning methods~\cite{liu2019multi} typically have one shared feature representation layer (e.g., Transformer) plus multiple task-specific prediction layers. It is unclear what types of knowledge or skills are learned in the feature representation layer. Unlike multi-task learning methods, SkillNet-NLU includes multiple skill modules with clear definitions. Skill modules are sparsely activated depending on the necessity to the task. 
Intuitively, SkillNet-NLU does not overspecialize at the task level, but at an inherent skill level through learning how each skill module works and how multiple skill modules are combined to tackle problems. We believe SkillNet-NLU generalizes better to new tasks with unforeseen task definitions in the future.
\item MoE methods typically include multiple homogeneous neural modules (called experts) in parallel, as given in Figure \ref{fig:model-intro} (d), and fully activate all the experts or partially activate a part of experts guided by an additional parameterized gating module~\cite{shazeer2017outrageously,lepikhin2020gshard,fedus2021switch,du2021glam}. However, what type of knowledge is learned in each expert is vague and why some experts are activated is not interpretable.\footnote{An exception is a recent work on machine translation where experts are selected based on the target language or language pair~\cite{kudugunta2021beyond}.} In SkillNet-NLU, the definition of each skill module is clear and the reason for a skill module being activated is that the skill is necessary (judged by human developers or users) to solve the task.
\end{enumerate}

We use Transformer~\cite{vaswani2017attention} and BERT~\cite{devlin2018bert} as the backbone to develop our system. 
Transformer is a commonly used model architecture with multiple layers and each layer is composed of a multi-head attention network followed by a feed-forward neural network (FFN). 
There are many different ways to implement SkillNet-NLU, and our goal is to demonstrate that a simple implementation works well in practice.
Specifically, we implement skill modules as homogeneous FFN networks.
A skill module is activated only if the skill is relevant to the task at hand.
Our model not only supports sparsely activated fine-tuning, but also can be pre-trained in the same sparse way through masked language modeling and next sentence prediction. 

We conduct experiments on Chinese natural language understanding tasks.
Experimental results on six tasks (including sentiment classification, natural language inference, semantic similarity, text classification, named entity recognition and machine reading comprehension) show that, with only one model checkpoint, our approach performs better than task-specific fine-tuning and two multi-task learning baselines:  a dense model and a Mixture-of-Experts model.
Furthermore, after being pre-trained with the same sparse manner, the overall performance is further boosted.
More importantly, we show that when being extended to new tasks, our approach significantly outperforms baseline systems.

\section{Background}
We give brief backgrounds on BERT and the standard BERT-based multi-task learning baseline.

BERT is a Transformer-based encoder~\cite{vaswani2017attention}. It is usually used in a pre-training and fine-tuning framework.
Model parameters are first pre-trained on a vast amount of unlabeled text data with self-supervised objectives (e.g., masked language modeling and next sentence prediction). 
Then, for each downstream task, the pre-trained model parameters are further fine-tuned on each task-specific data separately.
If there are $N$ downstream tasks, a standard solution would produce $N$ BERT models, each of which corresponds to a particular task. 

Since the smallest BERT model still has hundreds of millions of parameters, an efficient way of avoiding deploying multiple copies of big models in practice is to train one multi-task model to support multiple downstream tasks. 
A standard multi-task method~\cite{liu2019multi} appends different task-specific prediction layers on top of a shared Transformer layer.
In the training stage, all tasks are optimized jointly.
Intuitively, the Transformer layer learns the generic feature representations and each prediction layer learns to accomplish a particular task. 
In practice, conducting the second round of task-specific fine-tuning, namely fine-tuning model parameters for each task separately (i.e., producing $N$ models for $N$ tasks), might produce higher accuracy.
However, this contradicts to our motivation of developing one general-purpose model across multiple tasks. Therefore,  we don't conduct the second round of task-specific fine-tuning in our experiments.

\section{SkillNet-NLU}

This section gives our SkillNet-NLU and its application to natural language understanding tasks. 
We first describe the model architecture (\S \ref{sec:model}).
Then, we present the tasks used for model training (\S  \ref{sec:method-tasks}),  how to do multi-task training with SkillNet-NLU (\S \ref{sec:method-model-training}) and how to extend the model to new tasks (\S \ref{sec:method-new-tasks}).
Finally, we show how the model can be pre-trained with model parameters sparsely activated using traditional self-supervised learning objectives (i.e., masked language modeling and next sentence prediction) (\S \ref{sec:pre-training}).

\begin{table*}[t]
\centering
\resizebox{\textwidth}{!}{
\begin{tabular}{cccccccccc}
\toprule
\multirow{2}{*}{\bf Task Id} & \multirow{2}{*}{\bf Task} & \multicolumn{7}{c}{\bf Skills} & \multirow{2}{*}{\bf Dataset} \\
\cmidrule(lr){3-9}
& & \texttt{s1} & \texttt{s2} & \texttt{s3} & \texttt{s4} & \texttt{s5} & \texttt{s6} & \texttt{s7} & \\
\midrule
\texttt{T1} & Sentiment Analysis & \checkmark & & & \checkmark & & & \checkmark & ChnSentiCorp (9.6k / 1.2k) \\
\texttt{T2} & Natural Language Inference & \checkmark & & \checkmark & & & & \checkmark & OCNLI (50k / 3k)\\
\texttt{T3} & Semantic Similarity & \checkmark & & \checkmark & & & \checkmark & \checkmark & AFQMC (34.3k / 4.3k)\\
\texttt{T4} & Text Classification & \checkmark & & & & & & \checkmark & TNEWS (53.3k / 10k)\\
\texttt{T5} & Named Entity Recognition & & \checkmark & & & & & \checkmark & OntoNotes (15.7k / 4.3k)\\
\texttt{T6} & Machine Reading Comprehension & & \checkmark & \checkmark & & \checkmark & & \checkmark & CMRC 2018 (10k / 3.4k)\\
\bottomrule
\end{tabular}
}
\caption{Tasks and datasets used to train the multi-task model. Relevant skills (defined in Table \ref{table:intro-skills}) for each dataset is marked with a tick. The numbers of training and evaluation instances in each dataset are given in parentheses.}
\label{tab:task-skills}
\end{table*}

\subsection{Model Architecture}\label{sec:model}

There are many different ways to implement our SkillNet-NLU.
The goal of this work is to demonstrate that a simple and intuitive implementation of the idea works well in practice, and we leave the exploration of more advanced model architectures in the future.
Specifically, we build our SkillNet-NLU using Transformer~\cite{vaswani2017attention} and BERT~\cite{devlin2018bert} as the backbone.
Since both Transformer and BERT are ubiquitously adopted in natural language processing tasks, we don't elaborate on the details and 
refer readers to the original papers.

Transformer is a commonly used model architecture with multiple layers and each layer is composed of a multi-head attention network followed by a feed-forward neural network (FFN). 
Our model architecture modifies each of the Transformer layers and adds task-specific prediction layers on top of the representations of the last layer.

In Transformer, as given in Figure \ref{fig:model-architecture} (a),  each layer includes a multi-head attention network followed by a feed-forward neural network (FFN). 
In SkillNet-NLU, as shown in Figure \ref{fig:model-architecture} (b), we have a set of FFN layers in parallel, each of which stands for one particular skill (e.g., \verb|s1| from Table \ref{tab:task-skills}).
When being applied to one task, only the FFN layers corresponding to relevant skills are activated. For example, for the task of machine reading comprehension, only \verb|s2|, \verb|s3|, \verb|s5| and \verb|s7| are relevant, so the remaining FFN layers (i.e., \verb|s1|, \verb|s4| and \verb|s6|) are not activated.
Considering that the number of activated skills is variable, we accumulate the output vectors of activated skill FFN layers with average pooling.
The remaining operations are the same as the standard Transformer.

Specifically, given a sequence of input $x = \{x_1,...,x_n\}$,  our model first performs multi-head self-attention for each token.
Then, each skill module FNN$_k$ from the set of activated skills $S$ obtains skill-specific representations as follows,
\begin{equation}
\bm{h}_k = \text{FNN}_k(\text{Self-Attention}(\{x_1,...,x_n\})),
\end{equation}
where $k \in [1, |S|]$ indicates the $k$-th activated skill module in $S$.
For instance, for the task of machine reading comprehension, as shown in Figure \ref{fig:model-architecture} (c), $|S| = 4$ and $S = \{\verb|s2|, \verb|s3|, \verb|s5|, \verb|s7|\}$.
Finally, we adopt average-pooling over all the skill-specific representations to compute the output embeddings of words as follows,
\begin{equation}
\bm{v} = \text{AvgPool}(\bm{h}_1,...,\bm{h}_{|S|}).
\end{equation}
The aforementioned operations are performed for multiple rounds.
The embedding of each token produced by the last layer is considered as the final feature representation.

\subsection{Tasks}\label{sec:method-tasks}
We use six NLU tasks as given in Table \ref{tab:task-skills} to train our multi-task model.

\begin{figure*}[!t]
	\centering
	\subfigtopskip=2pt
	\subfigure[pre-training with masked language modeling. s2 and s7 activated.]{
	\includegraphics[width=0.47\linewidth]{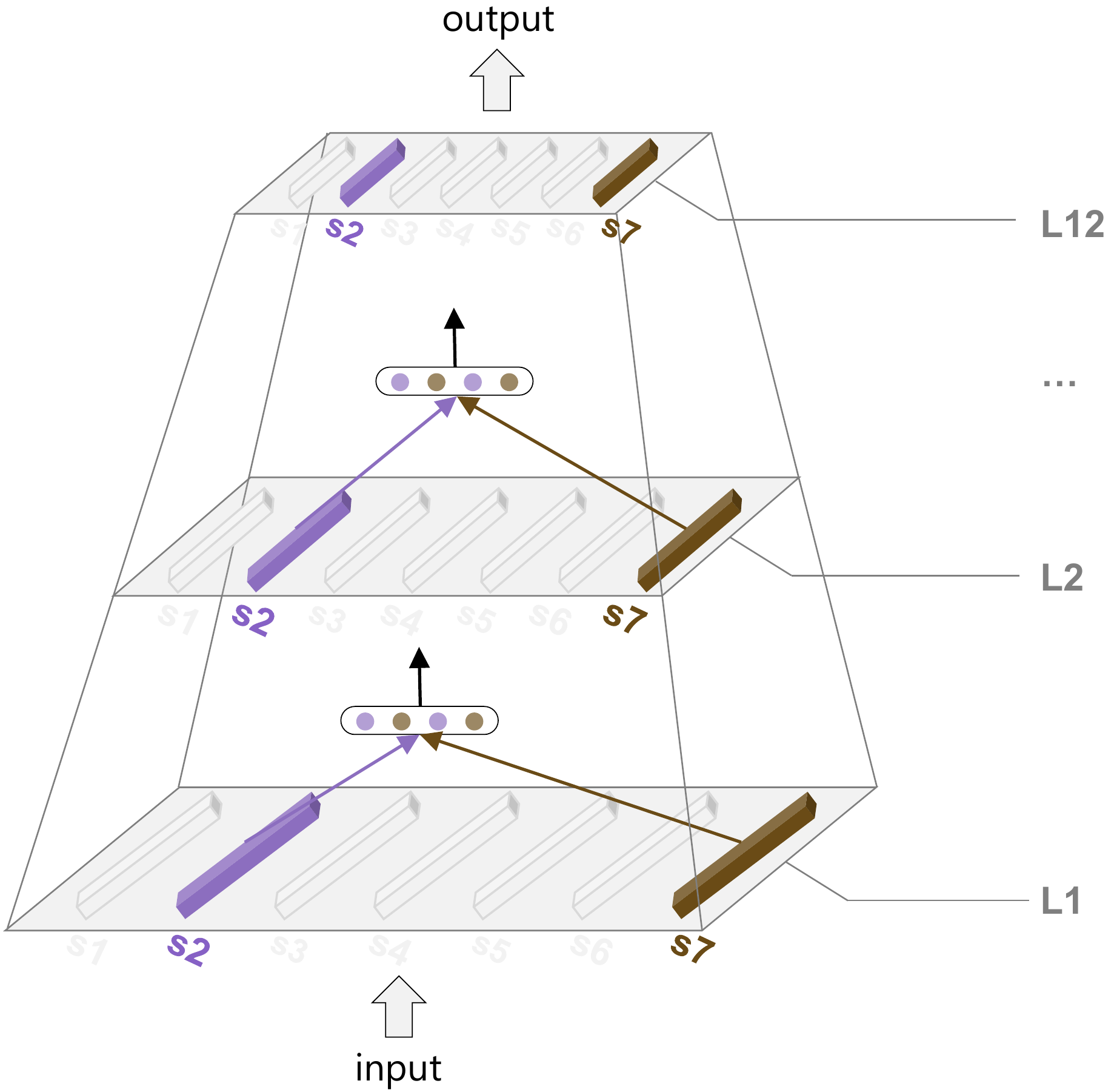}}\ \ \ \ \ \ \ \ 
	\subfigure[pre-training with next sentence prediction. s1, s3, s7 activated.]{
	\includegraphics[width=0.47\linewidth]{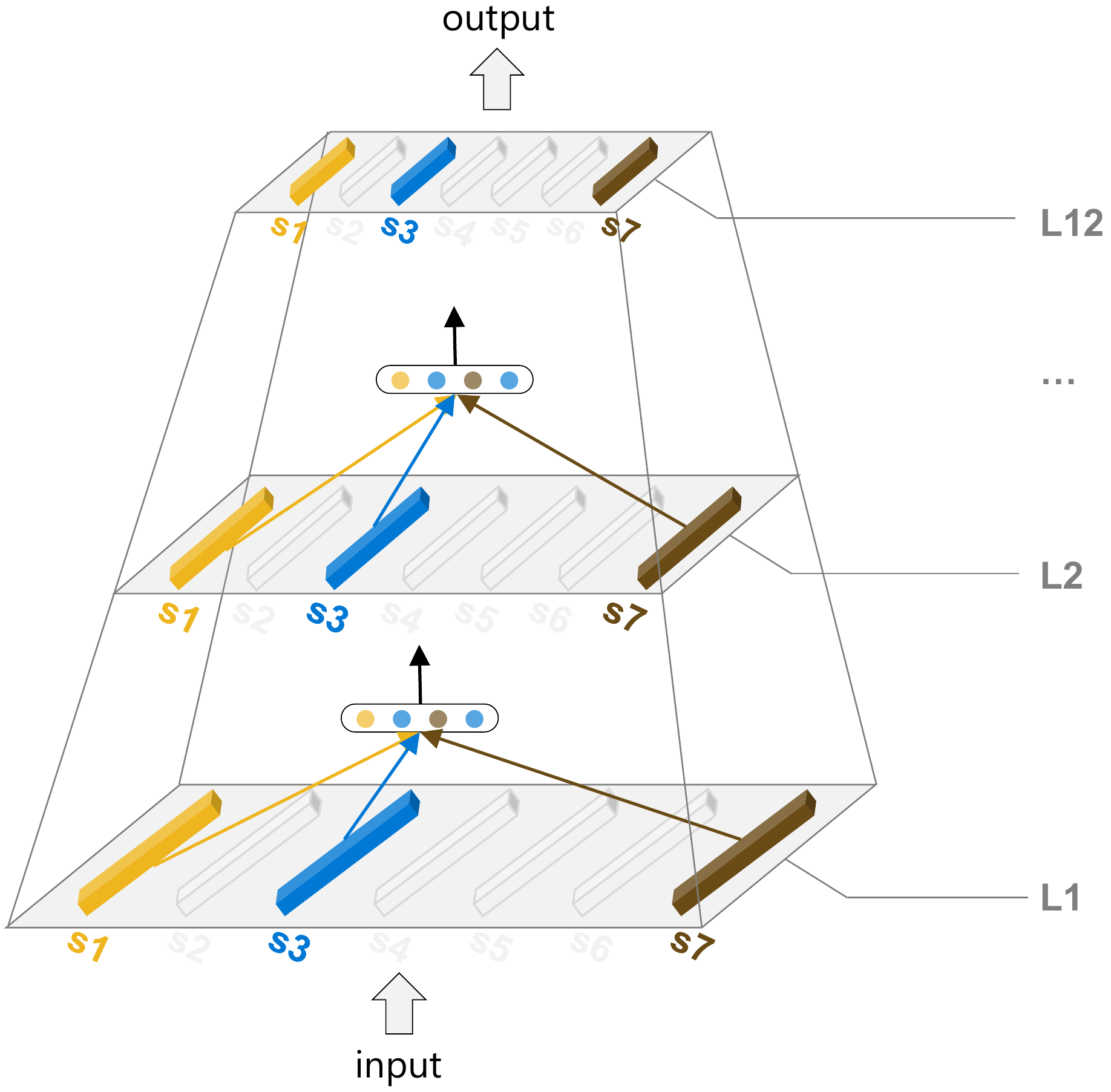}}
	\caption{An illustration of how our SkillNet-NLU is pre-trained with masked language modeling and next sentence prediction. The model is sparsely activated during pre-training. Skills are defined in Table \ref{table:intro-skills}.}
	\label{fig:model-pretrain}
\end{figure*}

\verb|T1| is sentiment classification. Given a text sequence (e.g., a sentence) as the input, the output is the polarity of the input. We consume the vector of $[\texttt{CLS}]$ to a softmax layer to conduct binary classification (i.e., positive v.s. negative).
\verb|T4| has the similar configuration.
We activate \texttt{s4} additionally for \verb|T2| because it requires the skill of understanding the sentiment in the texts.

\verb|T2| is natural language inference. Given two text sequences as the input, the output is the relation between two sequences as entailment, contradiction, or neutral. We concatenate two input segments with a $[\texttt{SEP}]$ token and consume the vector of $[\texttt{CLS}]$ to a softmax layer.
\verb|T3| has the analogous configuration. \texttt{s6} is activated in \verb|T3| because its data source comes from the finance domain.

\verb|T5| is named entity recognition. Given a sequence of words as the input, the task is detecting whether each word is a named entity, and if yes, predicting the entity type (e.g., person, organization, location, etc.).
We take the representations of each word from the last layer and feed them to Conditional Random Fields (CRF)~\cite{crf} to predict labels for words.

\verb|T6| is machine reading comprehension. Given a question and a passage as the input, the task is to predict a span from the passage that answers the question. 
The input of the model is the concatenation of the question and the passage, separated with a $[\texttt{SEP}]$ token. 
We take the representations of words from the passage and predict whether each of them is the starting index or the ending index of the answer.
Specifically, we introduce a start vector $v_{start}$ and an end vector $v_{end}$.
When predicting the probability of a token being the start of the answer span, we perform dot product between its vector and $v_{start}$ followed by softmax over all of the tokens in the paragraph.
The analogous formula is used for predicting the ending index.

\subsection{Model Training}\label{sec:method-model-training}
The overall training objective is to minimize the sum of the losses of all tasks.
Specifically, the model is trained on the concatenation of training samples from these tasks.
In each iteration, a mini-batch is selected from one task, and the model parameters are updated according to the task-specific objective.
We sample mini-batches from the $N = 6$ tasks according to a multinomial distribution with probabilities $\{q_i\}_{i=1...N}$:
\begin{equation}
q_{i} = \frac{p_{i}^{\alpha}}{\sum_{j=1}^{N} p_{j}^{\alpha}}\ \ \text{with}\ \ p_{i} = \frac{|T_{i}|}{\sum_{j=1}^{N}|T_{j}|},
\end{equation}
where $|T_{i}|$ indicates the number of training samples in task $T_i$.

The sampling rate $\alpha$ is a hyper-parameter to balance various tasks.
If $\alpha = 0.0$, $q_{i} = \frac{1}{N}$. Each task is selected by the equal chance.
Sampling with this distribution increases the number of samples associated with tasks with small size and alleviates the bias towards high-resource tasks.
If $\alpha = 1.0$,  the natural distribution of the tasks will be maintained and  low-resource tasks are not up-sampled.
We set the sampling rate $\alpha = 1.0$ in experiments.
Analysis on the influence of $\alpha$ is given in subsection~\ref{ana:alpha}.

\subsection{Adaptation to New Tasks}\label{sec:method-new-tasks}
We describe the adaptation of a well-trained multi-task SkillNet-NLU to new tasks. 
We consider two situations here, depending on whether new skills are required to tackle the new task.

The first situation is that existing skills considered in the multi-task training stage are sufficient to tackle the new task.
Consider the new task of open domain question answering that determines whether a sentence from the given documents answers the question. Despite exactly the same task is unseen in the training stage, the relevant skills (i.e., the skill of getting the semantic representation of a sequence (\texttt{s1}), the skill of understanding question (\texttt{s5}) and the skill of understanding how two segments interact (\texttt{s3})) are seen during multi-task training. Therefore, we use the standard framework that only activates relevant skills to tune model parameters for the new task.

The second situation is that the new task may need new skills that are unseen in the multi-task training stage. 
For example, the task of Chinese medical question-answer matching may require an additional skill of understanding texts in the medical domain, which is unseen in the multi-task training stage.
Our model supports two ways to learn for such new tasks.
One way is to keep the number of skills unchanged and, intuitively, learn the unseen skills (like medical text understanding) in the general skill (\texttt{s7}).
Another way is to add a new skill (\texttt{s8}), that is activated together with other activated skills to learn for the new task.

\begin{table*}[t]
\centering
\begin{tabular}{l|cccccc|c}
\toprule
 & \texttt{T1} & \texttt{T2} & \texttt{T3} & \texttt{T4} & \texttt{T5} & \texttt{T6} & \bf Avg \\
\midrule
BERT Fine-tuning & \bf 94.7$^{\dagger}$ & 74.6$^{\dagger}$ & \bf 74.2$^{\ddagger}$ & 56.1$^{\ddagger}$ & 78.2$^*$ & 84.5$^{\dagger}$ & 77.1 \\
\midrule
Task-specific fine-tuning & 94.3 & 75.0 & 72.3 & 56.9 & 79.2 & 84.8 & 77.1 \\
Joint fine-tuning (Dense) & 93.4 & 75.1 & 71.0 & \bf 57.4 & 78.2 & 83.8 & 76.5 \\
Joint fine-tuning (MoE) & 94.0 & 74.0 & 71.4 & 57.3 & 78.8 & 84.5 & 76.7 \\
\midrule
SkillNet-NLU w/o sparse pre-training & 94.1 & \bf 75.3 & 72.1 & 56.9 & 81.2 & 84.6 & 77.4 \\
SkillNet-NLU w/ sparse pre-training & 94.4 & 75.0 & 73.9 & 57.0 & \bf 81.5 & \bf 85.7 & \bf 77.9 \\
\bottomrule
\end{tabular}
\caption{Evaluation results on the six tasks during multi-task training. We report accuracy for \texttt{T1} $\sim$ \texttt{T4} and F1 for \texttt{T5} $\sim$ \texttt{T6}. \textbf{Avg} is the average score of all tasks. Results with $^{\dagger}$, $^{\ddagger}$ and $^*$ are based on google BERT from~\newcite{cui-etal-2021-pretrain}, ~\newcite{xu2020clue} and our experiments, respectively.}
\label{tab:main}
\end{table*}

\subsection{Sparse Pre-training}\label{sec:pre-training}
In this part, we show how the parameters of SkillNet-NLU can be pre-trained with model parameters being sparsely activated.
We adopt two standard self-supervised learning objectives~\cite{devlin2018bert} including masked language modeling (MLM) and next sentence prediction (NSP).
To be specific, we activate two skills $S_{MLM} = \{\verb|s2|, \verb|s7|\}$ for the MLM task. 
For the NSP task, three skills $S_{NSP} = \{\verb|s1|, \verb|s3|, \verb|s7|\}$ are activated.
We sampled the two tasks with the equal chance and the overall learning objective is to minimize the sum of the two losses.
We refer readers to~\newcite{devlin2018bert} for the details of these two pre-training tasks.
After being pre-trained, the parameters of pre-trained skills can be used to initialize the multi-task model.

\begin{table*}[t]
\centering
\begin{tabular}{lcccc}
\toprule
 & \bf \makecell[c]{\#Params  Activated} & \bf Dev & \bf Test \\
\midrule
BERT Fine-tuning$^{\dagger}$ & 102M & 80.7 & 80.8 \\
\midrule
Task-specific fine-tuning (BERT-base) & 102M & 80.3 & 80.9 \\
Task-specific fine-tuning (RoBERTa-large) & 326M & 82.7 & 83.2 \\
Joint fine-tuning (Dense) & 102M & 80.7 & 81.6 \\
Joint fine-tuning (MoE) & 159M & 81.0 & 82.4 \\
\midrule
SkillNet-NLU w/o sparse pre-training & 272M & 81.5 & 83.2 \\
SkillNet-NLU w/ sparse pre-training & 272M & \bf 83.9 & \bf 84.4 \\
\bottomrule
\end{tabular}
\caption{Evaluation results on the NLPCC-DBQA dataset. We report the F1 score on the dev and test set. Results with $^{\dagger}$ are based on google BERT from~\newcite{sun2019ernie}.}
\label{tab:dbqa}
\end{table*}

\section{Experiments}
This section is organized as follows.
We first describe experiment settings (\S \ref{sec:experiment-setup}), and then report results on multiple tasks (\S \ref{sec:experiment-results}).
Then, we present results on two new tasks (\S \ref{sec:experiment-new-tasks}).

\subsection{Experimental Setup}\label{sec:experiment-setup}

\paragraph{Datasets}
We conduct multi-task training on six Chinese natural language understanding datasets to evaluate the performance of the models.

{\bf ChnSentiCorp}~\cite{song6chnsenticorp} is a sentiment analysis dataset, where the text should be classified into either a positive or negative label.
{\bf OCNLI}~\cite{hu-etal-2020-ocnli} is a large-scale Chinese NLI dataset, which requires to predict the relation of premise-hypothesis pairs.
The labels contain contradiction, neutral and entailment.
{\bf AFQMC}~\cite{xu2020clue} is a binary classification dataset from the financial domain, which aims to predict whether two sentences are semantically similar.
{\bf TNEWS}~\cite{xu2020clue} is a short text classification dataset consisting of news titles, which requires to classify into one of 15 classes.
{\bf OntoNotes}~\cite{weischedel2013ontonotes} is designed for named entity recognition. The entities contain several types including person, organization and location, etc.
{\bf CMRC 2018}~\cite{2019} is a span-extraction machine reading comprehension dataset, which requires to extract a passage span for the given question.
Table~\ref{tab:task-skills} shows the detailed statistics of these datasets.

\paragraph{Baselines}
We compare our SkillNet-NLU with the following approaches:

$\bullet$ {\bf Task-specific fine-tuning}: We fine-tune all the parameters of our BERT model\footnote{We collect 800G pre-training data from web news and blog articles, and train a Chinese BERT-base model with a batch size of 10,240.} for each task individually.
Therefore, we have a total of six task-specific models in our experiments.

$\bullet$ {\bf Joint fine-tuning (Dense)}: We adopt our BERT as a shared model to obtain feature representation and then feed it to multiple task-specific prediction layers.
The parameters of the BERT model and all the top layers are learned jointly on the six tasks.

$\bullet$ {\bf Joint fine-tuning (MoE)}: We set the number of the FFNs in each layer as seven and activate the top-2 FFNs for each token, determined by a gating module.
The parameters of these FFNs are initialized with our BERT model and updated with the task-specific prediction layers.

We build our SkillNet-NLU using the implementation of BERT-base by HuggingFace’s Transformers~\cite{wolf2020huggingfaces}\footnote{\url{https://github.com/huggingface/transformers}}, which has 12 Transformer encoder layers, and 768 hidden state dimensions.
We have two configurations to do multi-task training.
The first setting (\textbf{w/o sparse pre-training}) is that all skill modules are initialized with FFN layers from our Chinese BERT.
The second setting (\textbf{w/ sparse pre-training}) is that we use the parameters after sparse pre-training to initialize the skills.
The details of sparse pre-training is shown in Appendix~\ref{app:pre-training}.

We conduct multi-task training for 50k steps with a maximum length of 512, a batch size of 8.
We use Adam~\cite{kingma2014adam} as the optimizer with $\beta_1=0.9$, $\beta_2=0.98$, $\epsilon=1e^{-6}$.
The learning rate is warmed up over the first 5k steps to a peak value of $2e^{-5}$, and then linearly decayed.
We show the learning curve of each task in Appendix~\ref{app:curves}.

\subsection{Results}\label{sec:experiment-results}
Table~\ref{tab:main} shows the evaluation results of the baseline systems as well as the proposed models on six tasks.
The two multi-task learning baselines (i.e., Joint fine-tuning (Dense) and Joint fine-tuning (MoE)) perform slightly worse than task-specific fine-tuning.
Our SkillNet-NLU without pre-training outperforms the baseline systems and achieves an average score of 77.4\%, demonstrating the effectiveness of the sparse activation.
The performance of the model with sparse pre-training is further improved to 77.9\%, which indicates that the skill modules are learned better after pre-training with the same sparse manner.

\begin{table*}[t]
\centering
\begin{tabular}{lcccc}
\toprule
 & \bf \makecell[c]{Update Old Skills} & \bf \makecell[c]{\#Params Activated} & \bf Dev & \bf Test \\
\midrule
BERT Fine-tuning$^{\dagger}$ & & 110M & 78.6 & 78.2 \\
\midrule
Task-specific fine-tuning (BERT-base) & & 102M & 78.4 & 78.1 \\
Task-specific fine-tuning (RoBERTa-large) & & 326M & 78.9 & 78.7 \\
Joint fine-tuning (Dense) & & 102M & 78.5 & 78.3 \\
Joint fine-tuning (MoE) & & 159M & 78.7 & 78.4 \\
\midrule
\textit{No New Skills} \\
SkillNet-NLU w/o sparse pre-training & Y & 272M & 78.8 & 78.6 \\
SkillNet-NLU w/ sparse pre-training & Y & 272M & 79.0 & 78.9 \\
\midrule
\textit{Injecting New Skills} \\
SkillNet-NLU w/o sparse pre-training & N & 57M & 77.8 & 77.1 \\
SkillNet-NLU w/ sparse pre-training & N & 57M & 78.6 & 78.2 \\
SkillNet-NLU w/o sparse pre-training & Y & 329M & 79.2 & 79.0 \\
SkillNet-NLU w/ sparse pre-training & Y & 329M & \bf 79.5 & \bf 79.3 \\
\bottomrule
\end{tabular}
\caption{Evaluation results on the cMed dataset. We report the top-1 accuracy on the dev and test set. Results with $^{\dagger}$ are based on google BERT from~\newcite{cui2020chinese}.}
\label{tab:cmed}
\end{table*}

\subsection{Results on New Tasks}\label{sec:experiment-new-tasks}
In this section, we present the adaptation of a well-trained multi-task SkillNet-NLU to new tasks.
Results are reported in two settings, depending on whether no new skills are required.

The first new task is open domain question answering.
Given a question and a candidate sentence, the task is determining whether the sentence answers the question.
We concatenate the question and the candidate sentence with a \texttt{[SEP]} token and consume the vector of \texttt{[CLS]} to a softmax layer to conduct binary classification.
In this setting, no new skills are not injected.
So we activate a set of four relevant skills $S_{NLPCC-DBQA} = \{\verb|s1|, \verb|s3|, \verb|s5|, \verb|s7|\}$ and fine-tune all the parameters of these skill modules for the new task.

We conduct experiments on the NLPCC-DBQA dataset~\cite{duan2016nlpcc}.
Table~\ref{tab:dbqa} shows the number of activated parameters and the F1 score of various models.
We can see that our final system, SkillNet-NLU with sparse pre-training, performs better than the RoBERTa-large\footnote{We adopt RoBERTa-wwm-ext-large, which is pre-trained on more pre-training data with the whole word mask strategy.} baseline with smaller number of activated parameters.

We consider the second new task of Chinese medical question-answer matching.
Given a question and a candidate answer set, models are required to select the most relevant answer.
The input of the model is the concatenation of the question and a candidate answer, separated with a \texttt{[SEP]} token.
We activate a set of four skills $S_{cMed} = \{\verb|s1|, \verb|s3|, \verb|s5|, \verb|s7|\}$ and take the representation of the \texttt{[CLS]} to compute similarity between the question and the candidate answer.
We explore whether to inject a new skill (\verb|s8|) of understanding texts from the medical domain, which is unseen in the multi-task training stage.
If the new skill is injected, we can initialize its parameters with the general skill (\verb|s7|).
Then, the parameters of four activated skills, as well as the new skill, are fine-tuned on the training data.

We conduct experiments on the cMedQA~\cite{zhang2017chinese} dataset.
Table~\ref{tab:cmed} shows the number of activated parameters and the top-1 accuracy of various models.
We show the model performance by not injecting new skills in the second block.
We can see that our SkillNet-NLU without pre-training outperforms the three baseline systems, achieving a top-1 accuracy of 78.6\%.
The third block shows the results by injecting a new skill.
We can see that the performance of SkillNet-NLU with or without sparse pre-training is improved consistently.
The underlying reason is that the number of parameters increased.
Surprisingly, we find that only updating the new skill can achieve strong performance.

\section{Ablation Study and Analysis}
Evaluation results show that our SkillNet-NLU outperforms task-specific fine-tuning and two multi-task learning baselines.
In this section, we conduct a detailed ablation study and experimental analyses to better understand the proposed method.
All the results are based on SkillNet-NLU without sparse pre-training, where all skill modules are initialized with FFN layers from our Chinese BERT.

\subsection{Ablation Study}\label{sec:ablation}

We perform an ablation study to explore the effects of each skill.
To be specific, we delete one of the seven skills in turn, and then activate other corresponding skills for each task.
The ablation results are presented in Table~\ref{tab:ablation}.
From each row of the table, we can see that the average score decrease when any skill is removed in SkillNet-NLU, demonstrating that all the skills defined are helpful for the multi-task training.
There is a significant drop when deleting the general skill \texttt{s7}, because it is shared by all tasks.
We can see that the task performance drops sharply when some closely related skills are removed, especially for the skill that is unique to the task (i.e., \texttt{s4} for \texttt{T1}, \texttt{s5} for \texttt{T6}, \texttt{s6} for \texttt{T3}).
We also find that removing \texttt{s2} significantly affects the performance on \texttt{T5} $\sim$ \texttt{T6} while doesn't hurt the accuracy on \texttt{T1} $\sim$ \texttt{T4}.
The reason is that \texttt{T1} $\sim$ \texttt{T4} are sequence prediction tasks and \texttt{T5} $\sim$ \texttt{T6} are token prediction tasks. 
Removing \texttt{s2} makes the model overspecializing to sequence prediction tasks, while is less versatile to other tasks that require token prediction ability.

\begin{table*}[h]
\centering
\begin{tabular}{c|cccccc|c}
\toprule
 & \texttt{T1} & \texttt{T2} & \texttt{T3} & \texttt{T4} & \texttt{T5} & \texttt{T6} & \bf Avg \\
\midrule
SkillNet-NLU & 94.08 & \bf 75.25 & \bf 72.13 & 56.94 & \bf 81.19 & \bf 84.64 & \bf 77.37 \\
\midrule
\ \ -- w/o \texttt{s1} & 94.06 & 74.08 & 70.44 & 56.57 & 80.65 & 84.12 & 76.65 \\ 
\ \ -- w/o \texttt{s2} & \bf 94.24 & 75.22 & 71.34 & \bf 57.11 & 78.82 & 83.55 & 76.71 \\ 
\ \ -- w/o \texttt{s3} & 93.50 & 74.07 & 71.62 & 57.07 & 79.84 & 83.72 & 76.64 \\ 
\ \ -- w/o \texttt{s4} & 93.42 & 74.87 & 72.06 & 56.99 & 78.70 & 84.08 & 76.69 \\ 
\ \ -- w/o \texttt{s5} & 94.15 & 74.75 & 71.66 & 57.08 & 78.84 & 83.61 & 76.68 \\ 
\ \ -- w/o \texttt{s6} & 93.43 & 73.63 & 71.28 & 56.87 & 80.86 & 84.23 & 76.72 \\ 
\ \ -- w/o \texttt{s7} & 94.04 & 74.85 & 71.99 & 56.30 & 78.14 & 84.22 & 76.59 \\ 
\bottomrule
\end{tabular}
\caption{Ablation results on the six tasks during multi-task training.}
\label{tab:ablation}
\end{table*}

\subsection{Influence of the Sampling Rate}
\label{ana:alpha}

\begin{figure}[h]
\centering
\includegraphics[scale=1.2]{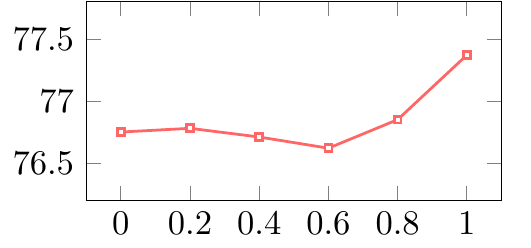}
\caption{Average score with different $\alpha$.}
\label{figure:labmda}
\end{figure}

As described in Section~\ref{sec:method-model-training}, we sample training examples from each task according to the sampling rate $\alpha$.
Figure \ref{figure:labmda} shows the average score with different $\alpha$.
We can see that the model performs better when the sampling rate $\alpha=1.0$, which maintains the natural distribution of the task.
The underlying reason is that the size of these datasets is relatively balanced.
The results also indicate that up-sampling datasets is consistently detrimental for multi-task learning, which is consistent with~\newcite{aghajanyan2021muppet}.
Therefore, we adopt $\alpha=1.0$ throughout all of our experiments.

\begin{table}[h]
\centering
\begin{tabular}{cc|c}
\toprule
\bf \#Num & \bf \#Params Total & \bf Avg \\
\midrule
3 & 187M & 76.5 \\
6 & 272M & 76.9 \\
9 & 357M & 77.2 \\
12 & 422M & 77.4 \\
\bottomrule
\end{tabular}
\caption{The number of total parameters and average score with the different number of top SkillNet-NLU layers.}
\label{tab:layer}
\end{table}

\subsection{Influence of the Number of Top SkillNet-NLU Layers}
\label{ana:layer}

We also investigate how the number of top SkillNet-NLU layers affects the model performance.
We conduct experiments based on SkillNet-NLU and the number of the top SkillNet-NLU layers varies from 3 to 12, increased by 3.
We show the number of total parameters and the average score of each model in Table~\ref{tab:layer}.
We can see that the performance consistently improves as the number grows, demonstrating the effectiveness of our SkillNet-NLU.
The underlying reason is that when more SkillNet-NLU layers are incorporated, the skills are better learned as the number of parameters increases.

\section{Conclusion}
In this work, we present a general-purpose model called SkillNet-NLU, and its application to natural language understanding tasks.
SkillNet-NLU includes a set of parameterized skill modules, and sparsely activate some of the modules depending on whether a skill is relevant to the target task.
The framework is generic and supports both multi-task fine-tuning and pre-training, both with sparse activation.
Results demonstrate that the approach performs better than baseline systems on both old and new tasks, and sparse pre-training brings further improvements.

This work can be further improved from many different angles, including defining a broader range of skills, exploring advanced model architectures, expanding from one language to multi languages or even from one modality to multiple modalities.


\bibliography{custom}
\bibliographystyle{acl_natbib}

\appendix

\section{Model Architecture}

\begin{figure*}[!t]
	\centering
	\subfigtopskip=2pt
	\subfigure[Each layer in Transformer]{
	\includegraphics[height=0.4\linewidth]{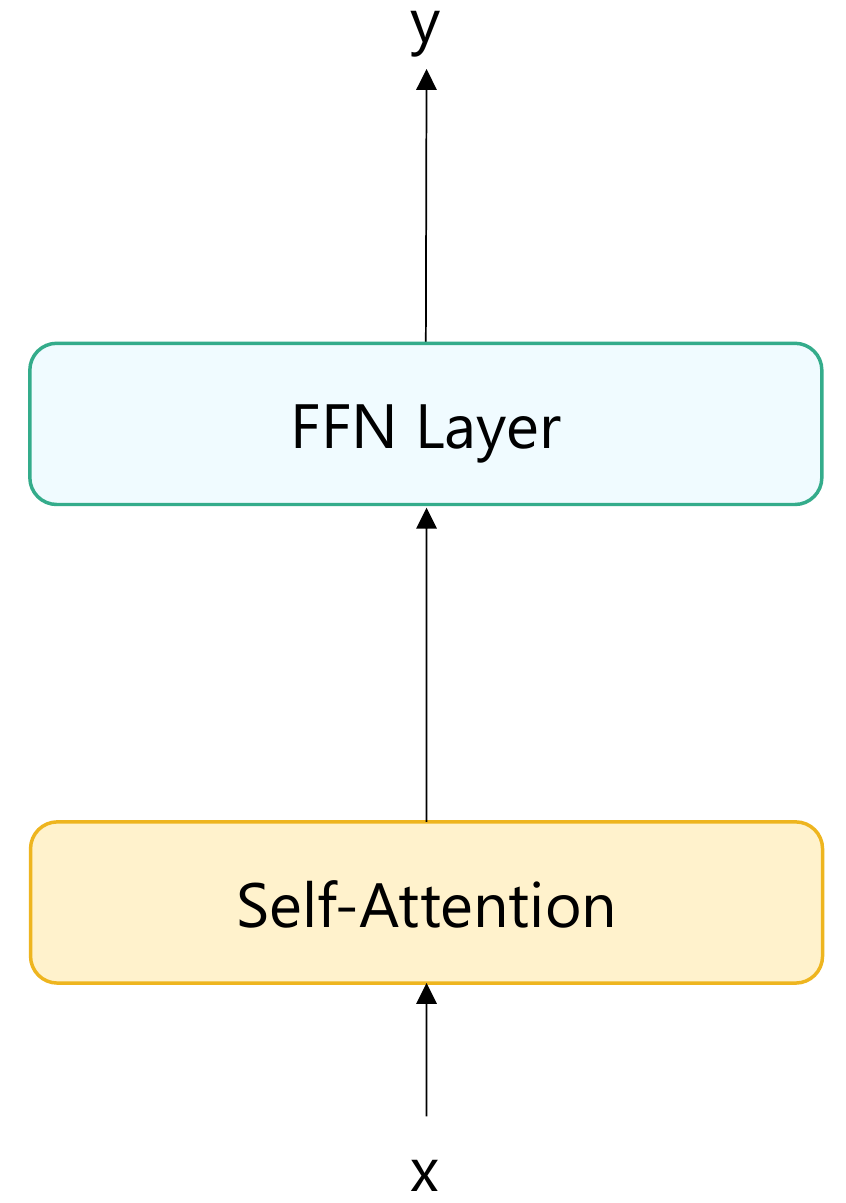}}\ \ \ \ \ \ \ \ 
	\subfigure[Each layer in SkillNet-NLU]{
	\includegraphics[height=0.4\linewidth]{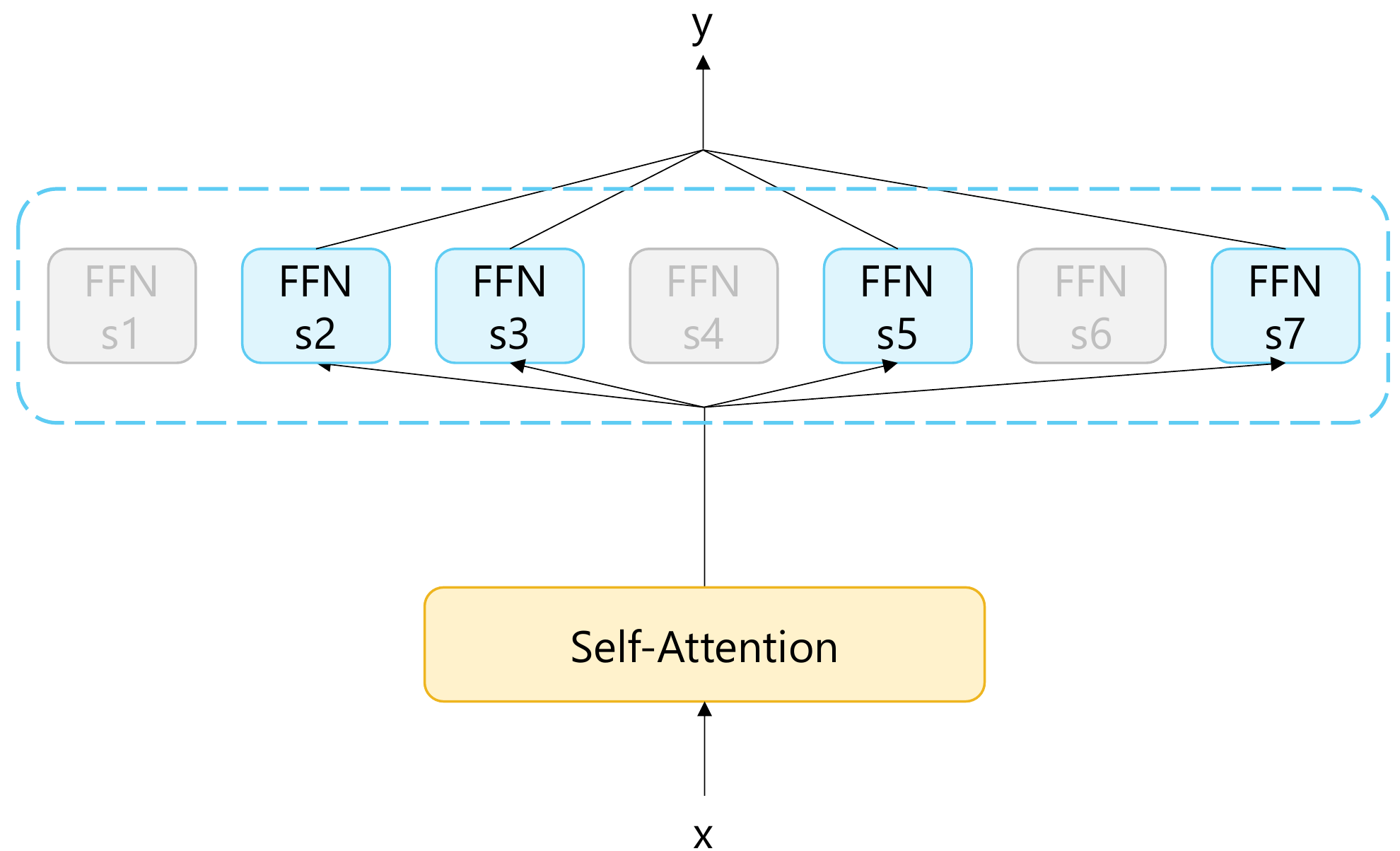}}
	\caption{A simple implementation of SkillNet-NLU (b) with comparison to the standard Transformer (a). This example illustrates the application of SkillNet-NLU to machine reading comprehension, where s2, s3, s5 and s7 are activated.}
	\label{fig:model-architecture}
\end{figure*}

\section{Sparse Pre-training Details}\label{app:pre-training}
During pre-training, we initialize four skill modules (i.e., \texttt{s1}, \texttt{s2}, \texttt{s3} and \texttt{s7}) with FFN layers from our Chinese BERT.
We adopt the same pre-training data and batch size that is used during the pre-training of our Chinese BERT.
SkillNet-NLU is pre-trained with mixed-precision training on 32 Nvidia Tesla V100 32GB GPUs for 100k steps with a maximum length of 512.
We use Adam~\cite{kingma2014adam} as the optimizer with $\beta_1=0.9$, $\beta_2=0.98$, $\epsilon=1e^{-6}$.
The learning rate is warmed up over the first 10k steps to a peak value of $3e^{-5}$, and then linearly decayed.

After being pre-trained, we can build a multi-task model by initializing the corresponding four skill modules.
The parameters of other three skill modules (i.e., \texttt{s4}, \texttt{s5} and \texttt{s6}) are initialized from the general skill module \texttt{s7}.

\section{Learning Curves}\label{app:curves}
\begin{figure*}[!t]
\centering
\includegraphics[scale=1.0]{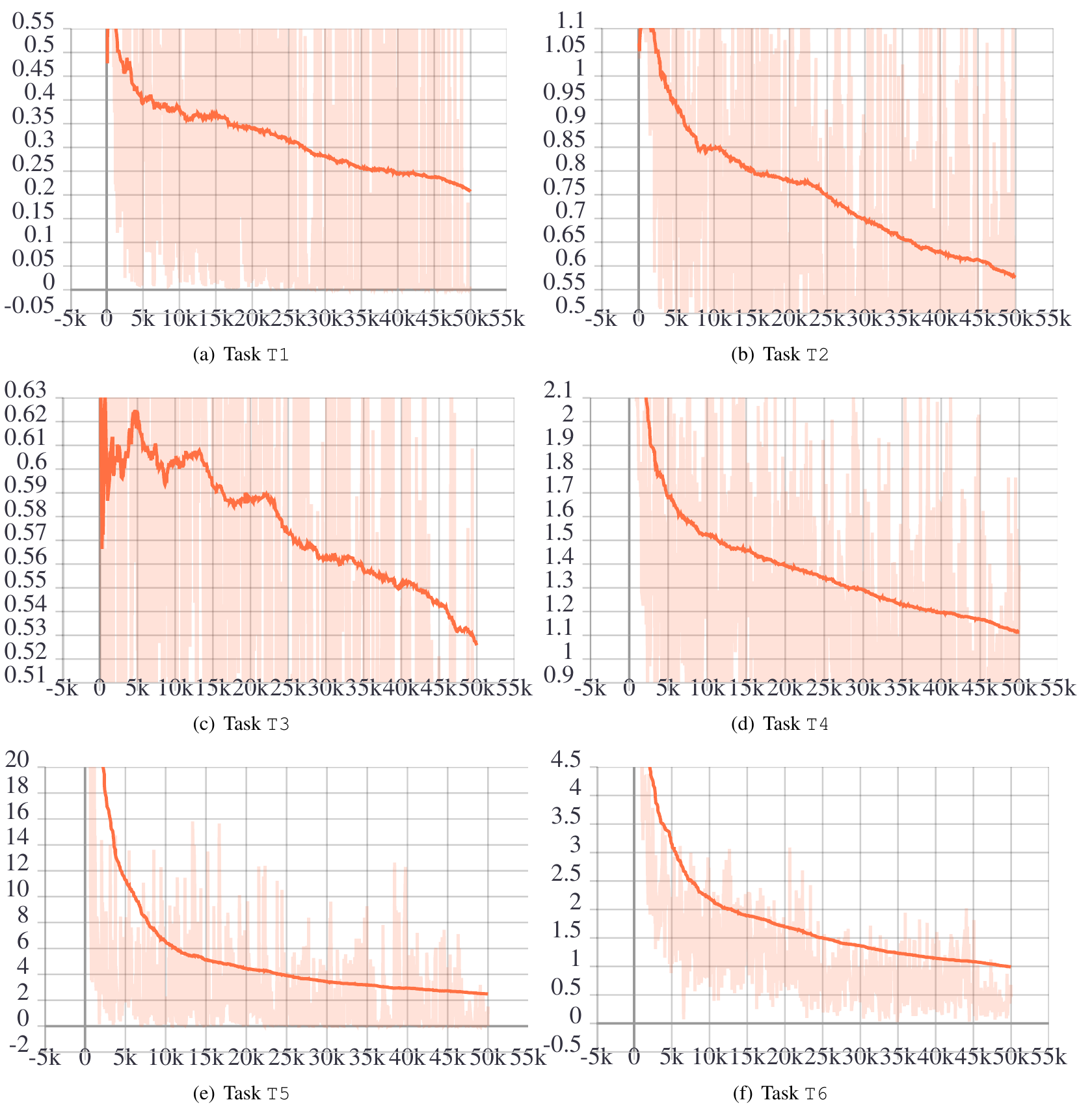}
\caption{The learning curve of each task during multi-task training.}
\label{figure:loss}
\end{figure*}

We show the learning curves during multi-task training in Figure~\ref{figure:loss}.

\end{CJK}
\end{document}